\pdfoutput=1
\documentclass[sigconf]{acmart}

\setcopyright{cc}
\setcctype{by}
\copyrightyear{2026}
\acmYear{2026}
\acmDOI{}
\acmISBN{}

\acmConference[L@S '26]{The 13th ACM Conference on Learning @ Scale}{June 29--July 3, 2026}{Seoul, South Korea}

\usepackage{tikz}
\usetikzlibrary{positioning, arrows.meta, calc, fit, backgrounds,
  shapes.geometric, patterns, decorations.pathreplacing}

\definecolor{cnnblue}{HTML}{4A90D9}
\definecolor{specgreen}{HTML}{6AB04C}
\definecolor{transyellow}{HTML}{F0C040}
\definecolor{ctcorange}{HTML}{E67E22}
\definecolor{agepurple}{HTML}{9B59B6}
\definecolor{lossred}{HTML}{E74C3C}
\definecolor{infergray}{HTML}{BDC3C7}
\definecolor{bgblue}{HTML}{EBF5FB}
\definecolor{bggray}{HTML}{F4F6F7}

\tikzset{
  B/.style={draw, rounded corners=2pt, minimum height=7mm,
    minimum width=18mm, align=center, font=\scriptsize\sffamily, line width=0.5pt},
  S/.style={draw, rounded corners=1.5pt, minimum height=6mm,
    minimum width=18mm, align=center, font=\tiny\sffamily, line width=0.4pt},
  H/.style={draw, rounded corners=2pt, minimum height=7mm,
    minimum width=18mm, align=center, font=\scriptsize\sffamily, line width=0.6pt},
  L/.style={draw=lossred, fill=lossred!8, rounded corners=2pt,
    minimum height=6mm, minimum width=14mm, align=center,
    font=\scriptsize\sffamily\bfseries, line width=0.6pt},
  D/.style={draw=black!50, fill=bggray, rounded corners=1.5pt,
    minimum height=6mm, align=center, font=\tiny\sffamily},
  a/.style={-{Stealth[length=1.8mm]}, line width=0.45pt},
  da/.style={-{Stealth[length=1.8mm]}, line width=0.45pt, dashed},
  d/.style={font=\fontsize{5}{6}\selectfont\sffamily, text=black!50},
}

\begin{document}

\title{Edge Phoneme Recognition for Children's Speech through Age-Aware Training}


\author{Matthew Arboleda}
\affiliation{%
  \institution{Occidental College}
  \city{Los Angeles}
  \state{CA}
  \country{USA}
}
\email{arboleda@oxy.edu}

\author{Ryan Arboleda}
\affiliation{%
  \institution{Occidental College}
  \city{Los Angeles}
  \state{CA}
  \country{USA}
}
\email{rarboleda@oxy.edu}

\author{Sophie Haak}
\affiliation{%
  \institution{Occidental College}
  \city{Los Angeles}
  \state{CA}
  \country{USA}
}
\email{haak@oxy.edu}

\author{Sam Hjelmeset}
\affiliation{%
  \institution{Occidental College}
  \city{Los Angeles}
  \state{CA}
  \country{USA}
}
\email{hjelmeset@oxy.edu}

\author{Andrew Franck}
\affiliation{%
  \institution{Occidental College}
  \city{Los Angeles}
  \state{CA}
  \country{USA}
}
\email{franck@oxy.edu}

\author{Bingrui Yang}
\affiliation{%
  \institution{Occidental College}
  \city{Los Angeles}
  \state{CA}
  \country{USA}
}
\email{byang@oxy.edu}

\author{Jose Bustamante Ortiz}
\affiliation{%
  \institution{Occidental College}
  \city{Los Angeles}
  \state{CA}
  \country{USA}
}
\email{bustamanteor@oxy.edu}

\author{Yuanrong Shen}
\affiliation{%
  \institution{Occidental College}
  \city{Los Angeles}
  \state{CA}
  \country{USA}
}
\email{sheny@oxy.edu }

\author{Joel Walsh}
\affiliation{%
  \institution{Occidental College}
  \city{Los Angeles}
  \state{CA}
  \country{USA}
}
\email{jwalsh2@oxy.edu}

\renewcommand{\shortauthors}{Arboleda et al.}

\begin{abstract}
  Detecting phonemes from children's speech has historically been difficult due to the scarcity of training data, and unique characteristics of children's speech. During a phoneme detection competition, we found that training a lightweight model to predict the age of the learner, as well as the phoneme sequence, enabled a 94M-parameter model to outperform WavLM Large models (317M) on the target DrivenData distribution, and fall within approximately 0.04 CER of competition ensembles with 90 times the parameters. This has enabled the creation of PhonemeTrainer, an application that can run on most modern cellular phones. This will ultimately enable better Automated Speech Recognition (ASR) and pronunciation helper apps for children's speech, with the privacy and compliance benefits that come with edge processing.
\end{abstract}



\begin{teaserfigure}
  \includegraphics[width=\textwidth]{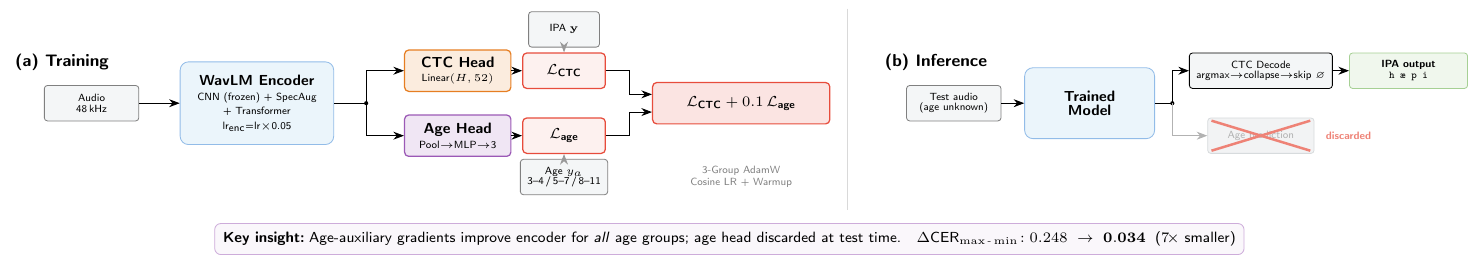}
  \caption{Multitask CTC with auxiliary age classification.}
  \Description{Training diagram showing WavLM encoder branching to CTC and age heads with combined loss, and
  simplified inference diagram showing trained model with age output discarded.}
  \label{fig:teaser}
\end{teaserfigure}

\raggedbottom
\maketitle

\section{Introduction}

Modern Automated Speech Recognition (ASR) and Speech to Phoneme pipelines are often built on Self-Supervised Learning representations trained on large amounts of adult speech \cite{li2024analysis, sinha2025beyond}. This results in representations that often fail at speech tasks that involve children's speech \cite{potamianos1997automatic, Dubagunta2019, Wang2026MindShift,BlockMedinetal2024}.

To this end, the Gates Foundation launched the ``On Top of Pasketti: Children's Speech Recognition Challenge'', a large-scale data science competition hosted on the DrivenData platform (concluded April 2026). \cite{drivendata2026pasketti, bull2016harnessing}. The challenge utilized a curated corpus of child speech derived from the Arizona Child Acoustic Database \cite{bunton2016arizona}, TalkBank \cite{rose2014phonbank, macwhinney2019understanding}, Jibo Kids \cite{shankar2024jibo}, and the ReadNet project \cite{readnet2024}. The phonetic track of this challenge required participants to build models that predict International Phonetic Alphabet phoneme sequences from children's speech utterances. The data included metadata about the speaker's age group buckets; ages 3--4, 5--7, 8--11, and 12+.

As with many data competitions like Kaggle, the top contest winners in ``On Top of Pasketti'' used ensemble models in order to achieve the lowest Character Error Rate (CER). The second place winner (0.2607 CER on the blind competition set) reported using an ensemble of 13 models, with each model utilizing encoders in the range of 317 million (WavLM Large) to 1.5 billion parameters (Whisper Large) \cite{dieleman2026pasketti}. The solution in this case provided negligible improvement in CER, with exponentially higher compute requirements (NVIDIA A100 GPU with $\sim$80 GB RAM). In order to create solutions that scale in primary/secondary school settings where cost is constrained, privacy is mandatory, and internet access is spotty, developers must have access to near state-of-the-art models that run on the edge.

\section{Training and Results}
 \begin{table}[h]                                                                                                                               
      \centering                                                                                                                                 
      \caption{Validation CER by Model Configuration. All runs used $N=117{,}500$. Per-age CERs are on the DD validation subset. (Age 3--4: 37,377 utterances, 5--7: 24,285 utterances, 8--11: 58,557 utterances).}                
      \label{tab:model-comparison}                                                                                                               
      \footnotesize                                                                                                                              
      \setlength{\tabcolsep}{2.5pt}                                                                                                              
      \begin{tabular}{@{}lccccccc@{}}                                                                                                            
      \toprule                                
      \textbf{Backbone} & \textbf{Dec.} & \textbf{DD} & \textbf{TB} & \textbf{Mix} & \textbf{3--4} & \textbf{5--7} & \textbf{8--11} \\           
      \midrule                                                                                                                                 
      Base+ (94M)          & CTC     & 0.649 & 0.664 & 0.656 & 0.658 & 0.628 & 0.706 \\                                                          
      \textbf{Base+ (94M)} & \textbf{CTC+age} & \textbf{0.306} & 0.326 & 0.316 & \textbf{0.323} & \textbf{0.325} & 0.290 \\
      Large (317M)         & CTC     & 0.360 & 0.356 & 0.358 & ---   & ---   & ---   \\                                                          
      Large (317M)         & RNN-T   & 0.343 & \textbf{0.280} & \textbf{0.306} & 0.334 & 0.412 & \textbf{0.162} \\                             
      \bottomrule                                                                                                                                
      \end{tabular}                                                                                                                              
\end{table}
The Age-aware training utilized a WavLM Base+ (94M parameters) \cite{chen2021wavlm} pretrained encoder and two decoder heads, one that predicts the age bucket, and one that uses Connectionist Temporal Classification \cite{graves2013speech} to predict International Phonetic Alphabet symbols (see Figure~\ref{fig:teaser}). Predicting age was not part of the competition criteria, but there was some hope that adding this head would result in the model learning a regularization signal. The age prediction was discarded for the purposes of the competition.

While training, this model converged three times faster than any other model we trained. All runs shared the same 117,500 utterance training set (10,696 DrivenData + 106,804 TalkBank), split by child ID to prevent data leakage. As Table~\ref{tab:model-comparison} shows, the 
  age-aware model was not merely competitive with WavLM Large (317M parameters) configurations---it matched the Large RNN-T on mixed CER (both 
  0.306) and outperformed it on the target DrivenData distribution (0.306 vs.\ 0.343), despite being 3.4$\times$ smaller. The WavLM Large + RNN-T model did perform better on TalkBank (TB) data, no doubt due to the     
  training data imbalance. On the actual competition DrivenData (DD), the age-aware model performed significantly better, neither
  under- nor overfitting on either subset of the data. This suggests that the age-auxiliary objective encourages the encoder to learn
  age-invariant phoneme representations rather than overfit on the more numerous (10:1 ratio) TalkBank data.

\section{Discussion}

Post-training tests of the Age-aware head showed that it was accurately predicting the age bucket of the speech samples with 72\% accuracy, which essentially meant predicting the dominant class every time. So while the head clearly sent a useful signal to the loss function, it was a subpar predictor. We conjecture that the age gradient may encourage age-invariant phoneme representations, preventing the encoder from overfitting to the acoustic characteristics of the majority age group. Future research will apply mechanistic interpretability techniques to explore the mechanisms by which this age-aware objective helps training and inference. 

The most important implication of this unique combination of size, accuracy, and quick convergence means that the model takes up a relatively small amount of device memory and can easily be trained and retrained as necessary to mitigate  distribution drift. This model can then serve as a stable backbone for a number of speech recognition and pronunciation feedback applications, without sending children's speech data over the internet. This has significant privacy implications in countries with strict regulations (laws like COPPA, FERPA, GDPR) governing the transmission of children's speech and academic records.
A key technical limitation arises when trying this with larger, more accurate models such as the finetuned WavLM-large (317M parameters) backbones. Those models often exceed 1 GB, creating challenges for mobile deployment including longer initial download times, RAM requirements that exceed Android's per-app memory limits, and inference delays that make real-time feedback impractical on mid-range devices.

\section{Demonstration}
\begin{figure}[h]
  \centering
  \includegraphics[width=0.25\linewidth]{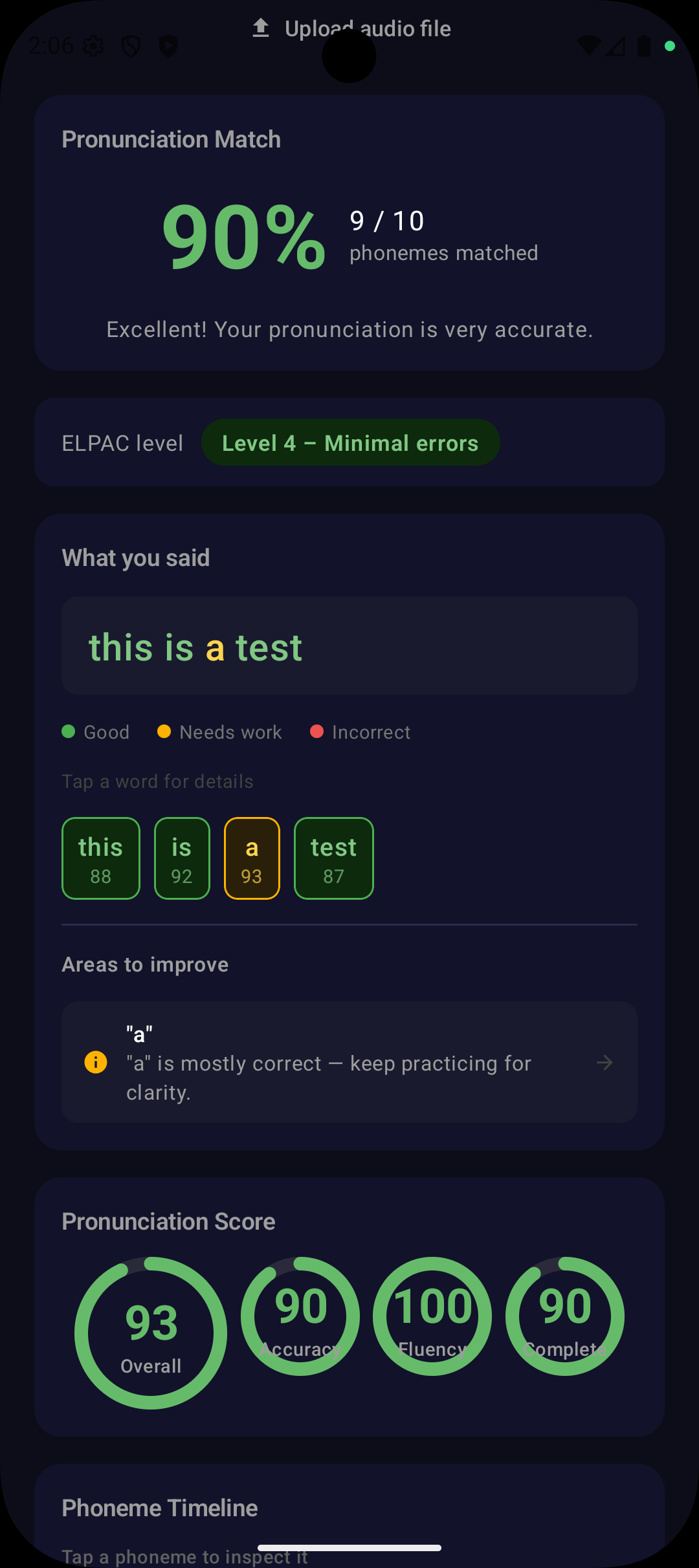}
  \caption{A compact view of the Android ELPAC UI.}
  \Description{Screenshot of the PhonemeTrainer Android app in dark mode, scrolled to show
  a completed pronunciation report. A ``Pronunciation Match'' card reports 90\%, 9 of 10
  phonemes matched, with the message ``Excellent! Your pronunciation is very accurate.''
  Below it, an ELPAC level badge reads ``Level 4 -- Minimal errors''. A ``What you said''
  card shows the transcribed sentence ``this is a test'' with each word colour-coded
  against a legend of green for good, amber for needs work, and red for incorrect; the
  word ``a'' is amber and the rest are green. Per-word score chips follow: this 88, is 92,
  a 93, test 87. An ``Areas to improve'' note explains that ``a'' is mostly correct and
  suggests practising for clarity. A ``Pronunciation Score'' card shows four ring gauges:
  93 overall, 90 accuracy, 100 fluency, and 90 completeness. A ``Phoneme Timeline''
  section begins at the bottom edge.}
  \label{fig:android_elpac_ui_small}
\end{figure}

This demonstration will show the phoneme transcription capabilities in real-time on an Android phone, as part of an application that gives users feedback on their pronunciation of short English, Spanish, or Korean sentences. During this demonstration participants will be asked to speak phrases into an Android device. These phrases will then be analyzed via an application called PhonemeTrainer. They will then be able to analyze the phonetic output of their speech, and analyze the accuracy to the reference phonetic sequence for that particular phrase (see Figure~\ref{fig:android_elpac_ui_small}). The application performs this analysis of participant speech by routing the .wav file to the finetuned WavLM Base+ acoustic model, which outputs a symbolic phoneme sequence. This spoken sequence is then mapped to ARPAbet phonemes and compared to the ground truth from the CMU Pronouncing Dictionary. Users are then shown a visual representation of their spoken sequence, and where, if at all, their phonetic output diverges from the ground truth sequence.

The demonstration video shows two audio samples, a three year old child and an adult each saying, ``this is a test'', to illustrate how the tool responds to different speaker profiles and voice characteristics. There will likely be no children in attendance. To show how the application performs on children's speech, we will pre-load several examples of children speaking the phrases for demonstration.

The participants will speak target sentences drawn from a language proficiency assessment used in many U.S. schools, selected to cover a representative range of phonemes. To manage the model's 360 MB size, the model weights are downloaded once from a remote host on first launch and saved to the device's internal storage, so that all subsequent sessions run fully offline without re-downloading. 

Future work will focus on using this speech-to-phoneme model as a complement to a language model that can take multilingual phrases and generate plausible phoneme sequences. This will allow PhonemeTrainer to work on any sentences, not just those that have been pre-loaded.

\bibliographystyle{ACM-Reference-Format}
\bibliography{references}


\begin{thebibliography}{16}


\ifx \showCODEN    \undefined \def \showCODEN     #1{\unskip}     \fi
\ifx \showISBNx    \undefined \def \showISBNx     #1{\unskip}     \fi
\ifx \showISBNxiii \undefined \def \showISBNxiii  #1{\unskip}     \fi
\ifx \showISSN     \undefined \def \showISSN      #1{\unskip}     \fi
\ifx \showLCCN     \undefined \def \showLCCN      #1{\unskip}     \fi
\ifx \shownote     \undefined \def \shownote      #1{#1}          \fi
\ifx \showarticletitle \undefined \def \showarticletitle #1{#1}   \fi
\ifx \showURL      \undefined \def \showURL       {\relax}        \fi
\providecommand\bibfield[2]{#2}
\providecommand\bibinfo[2]{#2}
\providecommand\natexlab[1]{#1}
\providecommand\showeprint[2][]{arXiv:#2}

\bibitem[Block~Medin et~al\mbox{.}(2024)]%
        {BlockMedinetal2024}
\bibfield{author}{\bibinfo{person}{Lucas Block~Medin}, \bibinfo{person}{Thomas
  Pellegrini}, {and} \bibinfo{person}{Lucile Gelin}.}
  \bibinfo{year}{2024}\natexlab{}.
\newblock \showarticletitle{Self-Supervised Models for Phoneme Recognition:
  Applications in Children's Speech for Reading Learning}. In
  \bibinfo{booktitle}{\emph{Proceedings of Interspeech 2024}}.
  \bibinfo{publisher}{ISCA}, \bibinfo{address}{Kos Island, Greece},
  \bibinfo{pages}{5168--5172}.
\newblock
\href{https://doi.org/10.21437/Interspeech.2024-1095}{doi:\nolinkurl{10.21437/Interspeech.2024-1095}}


\bibitem[Bull et~al\mbox{.}(2016)]%
        {bull2016harnessing}
\bibfield{author}{\bibinfo{person}{Peter Bull}, \bibinfo{person}{Isaac
  Slavitt}, {and} \bibinfo{person}{Greg Lipstein}.}
  \bibinfo{year}{2016}\natexlab{}.
\newblock \showarticletitle{Harnessing the power of the crowd to increase
  capacity for data science in the social sector}.
\newblock \bibinfo{journal}{\emph{arXiv preprint arXiv:1606.07781}}
  (\bibinfo{year}{2016}).
\newblock
\urldef\tempurl%
\url{https://arxiv.org/abs/1606.07781}
\showURL{%
\tempurl}


\bibitem[Bunton and Story(2016)]%
        {bunton2016arizona}
\bibfield{author}{\bibinfo{person}{Kate Bunton} {and} \bibinfo{person}{Brad~H.
  Story}.} \bibinfo{year}{2016}\natexlab{}.
\newblock \showarticletitle{Arizona Child Acoustic Database Repository}.
\newblock \bibinfo{journal}{\emph{Folia Phoniatrica et Logopaedica}}
  \bibinfo{volume}{68}, \bibinfo{number}{3} (\bibinfo{year}{2016}),
  \bibinfo{pages}{107--111}.
\newblock
\href{https://doi.org/10.1159/000452128}{doi:\nolinkurl{10.1159/000452128}}


\bibitem[Chen et~al\mbox{.}(2022)]%
        {chen2021wavlm}
\bibfield{author}{\bibinfo{person}{Sanyuan Chen}, \bibinfo{person}{Chengyi
  Wang}, \bibinfo{person}{Zhengyang Chen}, \bibinfo{person}{Yu Wu},
  \bibinfo{person}{Shujie Liu}, \bibinfo{person}{Zhuo Chen},
  \bibinfo{person}{Jinyu Li}, \bibinfo{person}{Naoyuki Kanda},
  \bibinfo{person}{Takuya Yoshioka}, \bibinfo{person}{Xiong Xiao},
  \bibinfo{person}{Jian Wu}, \bibinfo{person}{Long Zhou}, \bibinfo{person}{Shuo
  Ren}, \bibinfo{person}{Yanmin Qian}, \bibinfo{person}{Yao Qian},
  \bibinfo{person}{Jian Wu}, \bibinfo{person}{Michael Zeng},
  \bibinfo{person}{Xiangzhan Yu}, {and} \bibinfo{person}{Furu Wei}.}
  \bibinfo{year}{2022}\natexlab{}.
\newblock \showarticletitle{{WavLM}: Large-Scale Self-Supervised Pre-Training
  for Full Stack Speech Processing}.
\newblock \bibinfo{journal}{\emph{IEEE Journal of Selected Topics in Signal
  Processing}} \bibinfo{volume}{16}, \bibinfo{number}{6}
  (\bibinfo{year}{2022}), \bibinfo{pages}{1505--1518}.
\newblock
\href{https://doi.org/10.1109/JSTSP.2022.3188113}{doi:\nolinkurl{10.1109/JSTSP.2022.3188113}}


\bibitem[Dieleman(2026)]%
        {dieleman2026pasketti}
\bibfield{author}{\bibinfo{person}{Willem Dieleman}.}
  \bibinfo{year}{2026}\natexlab{}.
\newblock \bibinfo{title}{Now that we are done, who wants to talk about what
  worked?}
\newblock \bibinfo{howpublished}{DrivenData Community Forum}.
\newblock
\urldef\tempurl%
\url{https://community.drivendata.org/t/now-that-we-are-done-who-wants-to-talk-about-what-worked/11436/9}
\showURL{%
\tempurl}
\newblock
\shownote{Post \#9}.


\bibitem[{DrivenData}(2026)]%
        {drivendata2026pasketti}
\bibfield{author}{\bibinfo{person}{{DrivenData}}.}
  \bibinfo{year}{2026}\natexlab{}.
\newblock \bibinfo{title}{On Top of Pasketti: Children's Speech Recognition
  Challenge --- Phonetic Track}.
\newblock
  \bibinfo{howpublished}{\url{https://www.drivendata.org/competitions/309/childrens-phonetic-asr/}}.
\newblock
\newblock
\shownote{Accessed: 2026-04-08}.


\bibitem[Dubagunta et~al\mbox{.}(2019)]%
        {Dubagunta2019}
\bibfield{author}{\bibinfo{person}{S.~Pavankumar Dubagunta},
  \bibinfo{person}{Selen~Hande Kabil}, {and} \bibinfo{person}{Mathew
  Magimai-Doss}.} \bibinfo{year}{2019}\natexlab{}.
\newblock \showarticletitle{Improving Children Speech Recognition Through
  Feature Learning from Raw Speech Signal}. In
  \bibinfo{booktitle}{\emph{Proceedings of the 2019 IEEE International
  Conference on Acoustics, Speech and Signal Processing (ICASSP)}}.
  \bibinfo{publisher}{IEEE}, \bibinfo{address}{Brighton, UK},
  \bibinfo{pages}{5736--5740}.
\newblock
\href{https://doi.org/10.1109/ICASSP.2019.8682826}{doi:\nolinkurl{10.1109/ICASSP.2019.8682826}}


\bibitem[Graves et~al\mbox{.}(2013)]%
        {graves2013speech}
\bibfield{author}{\bibinfo{person}{Alex Graves}, \bibinfo{person}{Abdel-rahman
  Mohamed}, {and} \bibinfo{person}{Geoffrey Hinton}.}
  \bibinfo{year}{2013}\natexlab{}.
\newblock \showarticletitle{Speech recognition with deep recurrent neural
  networks}. In \bibinfo{booktitle}{\emph{Proceedings of the 2013 IEEE
  International Conference on Acoustics, Speech and Signal Processing
  (ICASSP)}}. \bibinfo{publisher}{IEEE}, \bibinfo{address}{Vancouver, BC,
  Canada}, \bibinfo{pages}{6645--6649}.
\newblock
\href{https://doi.org/10.1109/ICASSP.2013.6638947}{doi:\nolinkurl{10.1109/ICASSP.2013.6638947}}


\bibitem[Li et~al\mbox{.}(2024)]%
        {li2024analysis}
\bibfield{author}{\bibinfo{person}{Jialu Li}, \bibinfo{person}{Mark
  Hasegawa-Johnson}, {and} \bibinfo{person}{Nancy~L. McElwain}.}
  \bibinfo{year}{2024}\natexlab{}.
\newblock \showarticletitle{Analysis of Self-Supervised Speech Models on
  Children's Speech and Infant Vocalizations}. In
  \bibinfo{booktitle}{\emph{Proceedings of the 2024 IEEE International
  Conference on Acoustics, Speech, and Signal Processing Workshops (ICASSPW)}}.
  \bibinfo{publisher}{IEEE}, \bibinfo{address}{Seoul, Republic of Korea},
  \bibinfo{pages}{550--554}.
\newblock
\href{https://doi.org/10.1109/ICASSPW62465.2024.10626416}{doi:\nolinkurl{10.1109/ICASSPW62465.2024.10626416}}


\bibitem[MacWhinney(2019)]%
        {macwhinney2019understanding}
\bibfield{author}{\bibinfo{person}{Brian MacWhinney}.}
  \bibinfo{year}{2019}\natexlab{}.
\newblock \showarticletitle{Understanding spoken language through TalkBank}.
\newblock \bibinfo{journal}{\emph{Behavior Research Methods}}
  \bibinfo{volume}{51}, \bibinfo{number}{4} (\bibinfo{year}{2019}),
  \bibinfo{pages}{1919--1927}.
\newblock
\href{https://doi.org/10.3758/s13428-018-1174-9}{doi:\nolinkurl{10.3758/s13428-018-1174-9}}


\bibitem[{MIT Integrated Learning Initiative}(2024)]%
        {readnet2024}
\bibfield{author}{\bibinfo{person}{{MIT Integrated Learning Initiative}}.}
  \bibinfo{year}{2024}\natexlab{}.
\newblock \bibinfo{title}{ReadNet: Preventing Reading Failure with Speech
  Recognition-Powered Assessment}.
\newblock
  \bibinfo{howpublished}{\url{https://mitili.mit.edu/research/readnet-preventing-reading-failure-speech-recognition-powered-assessment}}.
\newblock
\newblock
\shownote{Accessed: 2026-04-08}.


\bibitem[Potamianos et~al\mbox{.}(1997)]%
        {potamianos1997automatic}
\bibfield{author}{\bibinfo{person}{Alexandros Potamianos},
  \bibinfo{person}{Shrikanth Narayanan}, {and} \bibinfo{person}{Sungbok Lee}.}
  \bibinfo{year}{1997}\natexlab{}.
\newblock \showarticletitle{Automatic speech recognition for children}. In
  \bibinfo{booktitle}{\emph{Proceedings of the 5th European Conference on
  Speech Communication and Technology (Eurospeech 1997)}}.
  \bibinfo{publisher}{ISCA}, \bibinfo{address}{Rhodes, Greece},
  \bibinfo{pages}{2371--2374}.
\newblock
\href{https://doi.org/10.21437/Eurospeech.1997-623}{doi:\nolinkurl{10.21437/Eurospeech.1997-623}}


\bibitem[Rose and MacWhinney(2014)]%
        {rose2014phonbank}
\bibfield{author}{\bibinfo{person}{Yvan Rose} {and} \bibinfo{person}{Brian
  MacWhinney}.} \bibinfo{year}{2014}\natexlab{}.
\newblock \showarticletitle{The PhonBank Project: Data and software-assisted
  methods for the study of phonology and phonological development}.
\newblock In \bibinfo{booktitle}{\emph{The Oxford Handbook of Corpus
  Phonology}}, \bibfield{editor}{\bibinfo{person}{Jacques Durand},
  \bibinfo{person}{Ulrike Gut}, {and} \bibinfo{person}{Gjert Kristoffersen}}
  (Eds.). \bibinfo{publisher}{Oxford University Press},
  \bibinfo{address}{Oxford, UK}, \bibinfo{pages}{380--401}.
\newblock


\bibitem[Shankar et~al\mbox{.}(2024)]%
        {shankar2024jibo}
\bibfield{author}{\bibinfo{person}{Natarajan~Balaji Shankar},
  \bibinfo{person}{Amber Afshan}, \bibinfo{person}{Alexander Johnson},
  \bibinfo{person}{Aurosweta Mahapatra}, \bibinfo{person}{Alejandra Martin},
  \bibinfo{person}{Haolun Ni}, \bibinfo{person}{Hae~Won Park},
  \bibinfo{person}{Marlen~Quintero Perez}, \bibinfo{person}{Gary Yeung},
  \bibinfo{person}{Alison Bailey}, \bibinfo{person}{Cynthia Breazeal}, {and}
  \bibinfo{person}{Abeer Alwan}.} \bibinfo{year}{2024}\natexlab{}.
\newblock \showarticletitle{The {JIBO} Kids Corpus: A speech dataset of
  child-robot interactions in a classroom environment}.
\newblock \bibinfo{journal}{\emph{JASA Express Letters}} \bibinfo{volume}{4},
  \bibinfo{number}{11} (\bibinfo{year}{2024}), \bibinfo{pages}{115201}.
\newblock
\href{https://doi.org/10.1121/10.0034195}{doi:\nolinkurl{10.1121/10.0034195}}


\bibitem[Sinha et~al\mbox{.}(2025)]%
        {sinha2025beyond}
\bibfield{author}{\bibinfo{person}{Abhijit Sinha},
  \bibinfo{person}{Hemant~Kumar Kathania}, {and} \bibinfo{person}{Mikko
  Kurimo}.} \bibinfo{year}{2025}\natexlab{}.
\newblock \showarticletitle{Beyond Traditional Speech Modifications: Utilizing
  Self Supervised Features for Enhanced Zero-Shot Children {ASR}}. In
  \bibinfo{booktitle}{\emph{Proceedings of Interspeech 2025}}.
  \bibinfo{publisher}{ISCA}, \bibinfo{address}{Rotterdam, The Netherlands},
  \bibinfo{pages}{1963--1967}.
\newblock
\href{https://doi.org/10.21437/Interspeech.2025-1874}{doi:\nolinkurl{10.21437/Interspeech.2025-1874}}


\bibitem[Wang et~al\mbox{.}(2026)]%
        {Wang2026MindShift}
\bibfield{author}{\bibinfo{person}{Zilai Wang},
  \bibinfo{person}{Natarajan~Balaji Shankar}, \bibinfo{person}{Kaiyuan Zhang},
  \bibinfo{person}{Zihan Wang}, {and} \bibinfo{person}{Abeer Alwan}.}
  \bibinfo{year}{2026}\natexlab{}.
\newblock \showarticletitle{Mind the Shift: Using Delta {SSL} Embeddings to
  Enhance Child {ASR}}.
\newblock \bibinfo{journal}{\emph{arXiv preprint arXiv:2601.20142}}
  (\bibinfo{year}{2026}).
\newblock
\urldef\tempurl%
\url{https://arxiv.org/abs/2601.20142}
\showURL{%
\tempurl}
\newblock
\shownote{Accepted to ICASSP 2026}.


\end{thebibliography}
\end{document}